\documentclass[10pt,twocolumn,letterpaper]{article}

\usepackage{cvpr}
\usepackage{times}
\usepackage{epsfig}
\usepackage{graphicx}
\usepackage{amsmath}
\usepackage{amssymb}

\usepackage{subcaption}
\usepackage{adjustbox}
\usepackage{booktabs}
\usepackage{url}
\usepackage{bbm}
\usepackage{amsmath}

\usepackage[pagebackref=true,breaklinks=true,letterpaper=true,colorlinks,bookmarks=false]{hyperref}

\cvprfinalcopy % *** Uncomment this line for the final submission

\def\cvprPaperID{5181} % *** Enter the CVPR Paper ID here

\ifcvprfinal\fi
\begin{document}

%%%%%%%%% TITLE
\title{Transfer Learning from Synthetic to Real-Noise Denoising with Adaptive Instance Normalization}
	
\author{Yoonsik Kim \and Jae Woong Soh  \and Gu Yong Park \and Nam Ik Cho \and
	\\Departmet of ECE, INMC, Seoul National University, Seoul Korea\\
	{\tt\small \{terryoo, soh90815, benkay\}@ispl.snu.ac.kr, nicho@snu.ac.kr} }
	

\maketitle

%\author{Yoonsik Kim	
%	% For a paper whose authors are all at the same institution,
%	% omit the following lines up until the closing ``}''.
%	% Additional authors and addresses can be added with ``\and'',
%	% just like the second author.
%	% To save space, use either the email address or home page, not both
%	\and
%	Jae Woong Soh	
%	\and
%	Gu Yong Park	
%	\and
%	Nam Ik Cho	
%	\\	
%	Departmet of ECE, INMC, Seoul National University, Seoul Korea\\
%	{\tt\small terryoo@ispl.snu.ac.kr}
%}
\maketitle
\thispagestyle{empty}

%%%%%%%%% ABSTRACT
\begin{abstract}
Real-noise denoising is a challenging task because the statistics of real-noise do not follow the normal distribution, and they are also spatially and temporally changing.
In order to cope with various and complex real-noise, we propose a well-generalized denoising architecture and a transfer learning scheme.
Specifically, we adopt an adaptive instance normalization to build a denoiser, which can regularize the feature map and prevent the network from overfitting to the training set.
We also introduce a transfer learning scheme that transfers knowledge learned from synthetic-noise data to the real-noise denoiser.
From the proposed transfer learning, the synthetic-noise denoiser can learn general features from various synthetic-noise data, and the real-noise denoiser can learn the real-noise characteristics from real data.
From the experiments, we find that the proposed denoising method has great generalization ability, such that our network trained with synthetic-noise achieves the best performance for Darmstadt Noise Dataset (DND) among the methods from published papers.
We can also see that the proposed transfer learning scheme robustly works for real-noise images through the learning with a very small number of labeled data. 

\end{abstract}

%%%%%%%%% BODY TEXT
\section{Introduction}
Image restoration tasks \cite{ gharbi2016deep, ledig2017photo, lehtinen2018noise2noise, lefkimmiatis2018universal, zhang2018learning, xiao2018discriminative, kokkinos2019iterative, manakov2019noise} have achieved noticeable improvement with the development of convolutional neural network (CNN).
Although most of image restoration methods work well on synthetically degraded images \cite{kim2016accurate, zhang2018image,cai2019toward,kim2019adaptively}, they show insufficient performance on the real degradations.

\begin{figure*}[t]
	\centering	
	\begin{subfigure}[b]{0.29\textwidth}		
		\includegraphics[width=1\columnwidth]{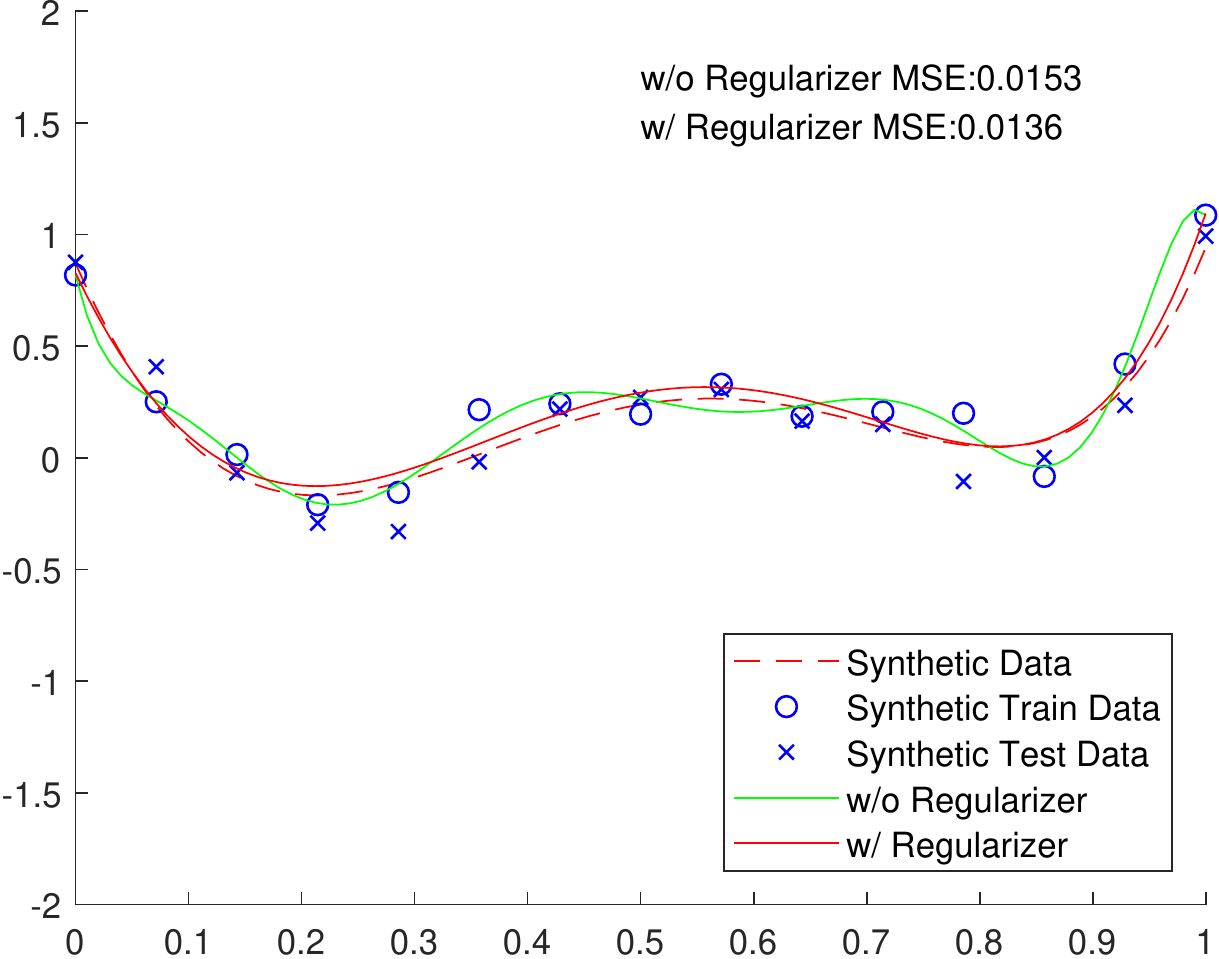} 
		\caption{Synthetic Data}		
	\end{subfigure}	
	\begin{subfigure}[b]{0.29\textwidth}		
		\includegraphics[width=1\columnwidth]{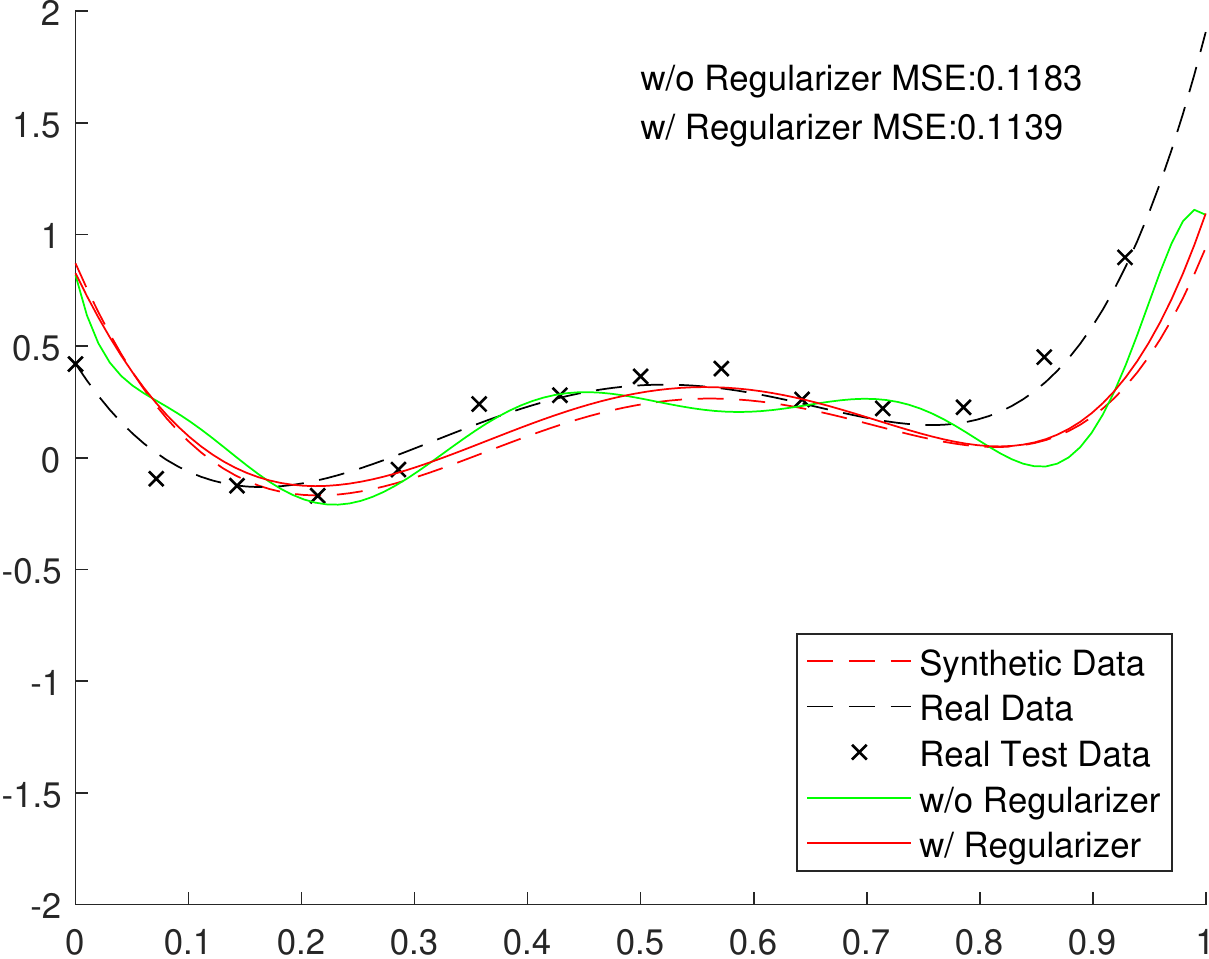} 
		\caption{Real Data}		
	\end{subfigure}	
	\begin{subfigure}[b]{0.29\textwidth}		
		\includegraphics[width=1\columnwidth]{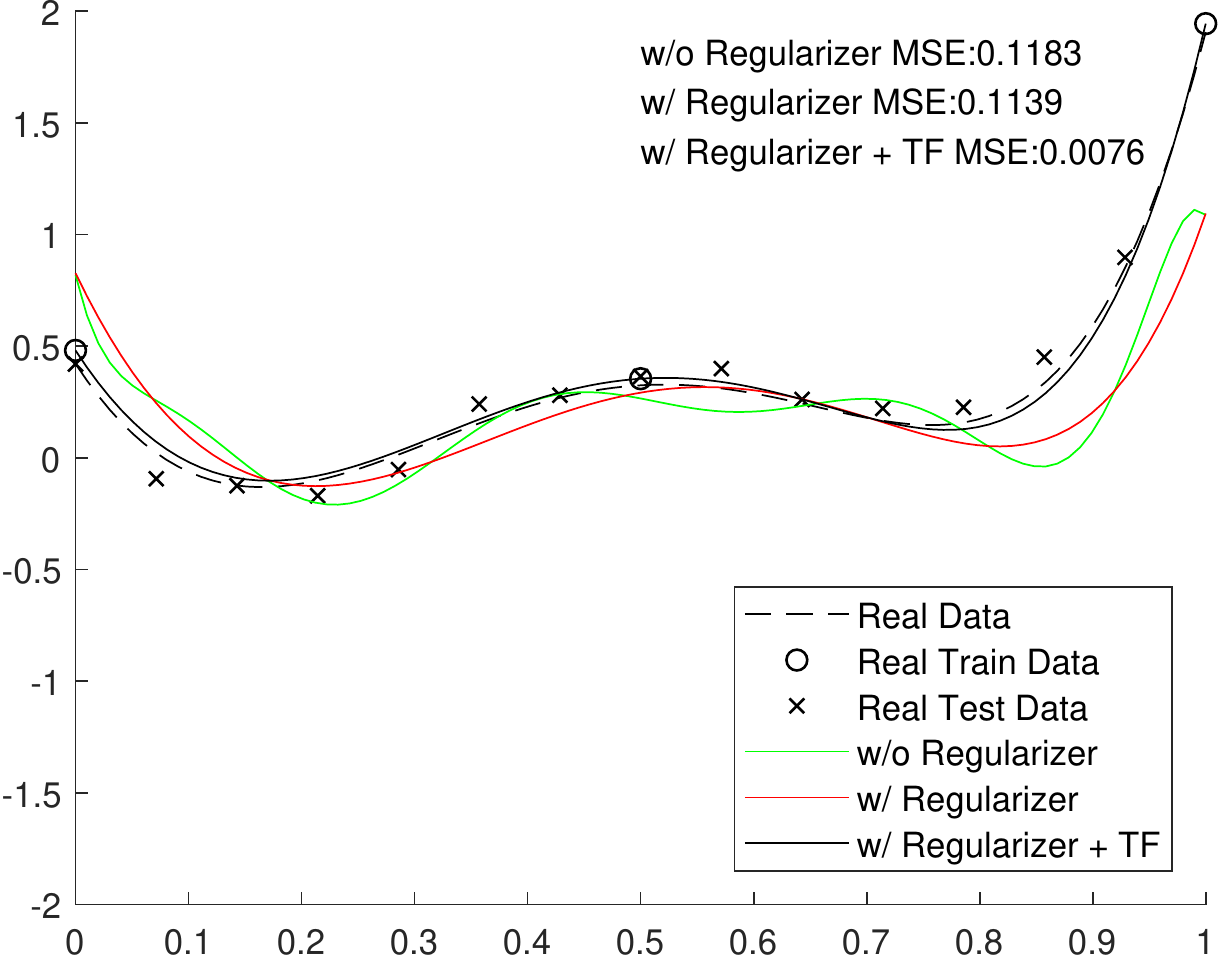} 
		\caption{Transfer Learning}		
	\end{subfigure}	
	\caption{
			A toy regression example presenting the effects of regularization and transfer learning.
		(a) We assume that training and test data are sampled from a 5th order polynomial, with additive white Gaussian noise (AWGN).
The original regression model (without regularizer) is denoted as \textit{w/o Regularizer}, which is a 10th order polynomial model. As well know, the higher-order model overfits the data. Assuming that a regularization method successfully degenerates the model to a 6th order one (\textit{w/ Regularizer}), then overfitting is relieved.
It can be seen from mean squared error (MSE) on synthetic test data that the regularization can enhance the performance when training and test distributions are the same.
		(b) 
We assume another 5th-order polynomial that generates a real data that has some domain difference from the synthetic one.
It can be seen from the MSE on real test data that the regularization is essential for processing other distributions.
		(c) Transfer learning regression method \textit{w/ Regularizer + TF} is fine-tuned with few real data samples from \textit{w/ Regularizer}.
		It can be seen from the MSE on real test data that transfer learning can be trained efficiently with few real training samples.
	}
	\label{fig:introduction}	
\end{figure*}

Regarding the denoising methods, the networks trained with synthetic-noise (SN) do not work well for the real-world images because of the discrepancy in the distribution of SN and real-noise (RN).
Specifically, CNNs~\cite{zhang2017beyond, zhang2017learning, zhang2018ffdnet} trained with Gaussian noise do not work well for the real-world images, because the CNNs are overfitted to the Gaussian distribution. The problem of overfitting can also be seen from a toy regression example in Fig.~\ref{fig:introduction}.
As shown in Fig.~\ref{fig:introduction}(a), the severely overfitted regression method (\textit{`w/o Regularizer'}) shows worse performance than a regularized method (\textit{`w/ Regularizer'}) on the synthetic test data.
Moreover, it can be seen in Fig.~\ref{fig:introduction}(b) that the generalization ability is much worse when the training and test domains are different.

To better address the problem due to the different data distribution between training and test sets, two kinds of approaches have been developed: (1) obtaining the pairs of RN image and corresponding near-noise-free image \cite{nam2016holistic,plotz2017benchmarking,anaya2018renoir,abdelhamed2018high,xu2018real}, and (2) finding more realistic noise model \cite{guo2019toward,brooks2019unprocessing}.

The RN datasets enable the quantitative comparison of denoising performance on real-world images and also provide the training sets for learning-based methods.
The CNNs trained with RN datasets robustly work on the real-world images, because domains of training and test set almost coincide.
However, acquiring the pairs of RN images needs specialized knowledge, and the amount of provided datasets would not be enough for training a deeper CNN~\cite{xu2017multi,xu2018external}.
Furthermore, learning-based methods can be easily overfitted to a specific camera device (dataset), which cannot cover all the devices that have different characteristics such as gamma correction, color correction, and other in-camera pipelines.

For a finding more realistic noise model,
CBDNet~\cite{guo2019toward} synthesized near-RN images by considering realistic noise models and simulating the in-camera pipeline.
It generates enough dataset that simulates more than 200 camera response functions.
The CBDNet shows excellent performance on RN images even though the CNN is trained with the SN. Furthermore, they showed that additional training with RN dataset improves performance.
Although realistic noise modeling indeed reduces the domain discrepancy between SN and RN, there still remains a domain discrepancy to be handled.
Moreover, CNN can be overfitted to a certain noise model that is actually not a `real' noise.

From these observations, we propose a novel denoiser that is well generalized to the various RN from camera devices by employing an adaptive instance normalization (AIN) \cite{ulyanov2016instance,huang2017arbitrary, li2017universal, park2019semantic}.
In recent CNN based methods for restoring the synthetic degradations \cite{lim2017enhanced, zhang2018image, kim2019adaptively}, regularization methods have not been exploited due to the small performance gain (even degrading performance).
This indicates that a CNN is overfitted to the training data to get the best performance when domains of training and test set coincide \cite{feng2019suppressing}.

On the other hand, the denoiser trained with SN needs regularization, for applying it to the RN denoising. As shown in the example of Fig.~\ref{fig:introduction} (a) and (b) with \textit{`w/ Regularizer'}, the network needs to be generalized through the regularization.
In this respect, we propose a well-regularized denoiser by adopting the AIN as a regularization method.
Specifically, the affine transform parameters for the normalization of features are generated from the pixel-wise noise level. Then, the transform parameters adaptively scale and shift the feature maps according to the noise characteristics, which results in the generalization of the CNN.

Furthermore, we propose a transfer learning scheme from SN to RN denoising network to reduce the domain discrepancy between the synthetic and the real.
As mentioned above, the RN dataset would not be sufficient to train a CNN, which can also be easily overfitted to a certain RN dataset.
Hence, we devise a transfer learning scheme that learns the general and invariant information of denoising from the SN domain and then transfer-learns the domain-specific information from the information of RN.
As can be seen in Fig.~\ref{fig:introduction}(c), we believe that the SN denoiser can be adapted to an RN denoiser by re-transforming normalized features.
Specifically, the parameters of AIN are updated using the RN dataset. The proposed scheme based on transfer learning can be applied to any dataset that has a small number of labeled data. That is, a CNN trained with the SN is easily transferred to work for the RN removal, without the need for training the whole network with the RN.

The contribution of this work can be summarized as follows:
\begin{itemize}
	\item We propose a novel well-generalized denoiser based on the AIN, which enables the CNN to work for various noise from many camera devices.	
	\item We introduce a transfer learning for the denoising scheme, which learns the domain-invariant information from SN data and updates affine transform parameters of AIN for the different-domain data.
	\item The proposed method achieves state-of-the-art performance on the SN and RN images.	
\end{itemize}
\begin{figure*}[t]
	\centering				
	\includegraphics[width=0.80\textwidth]{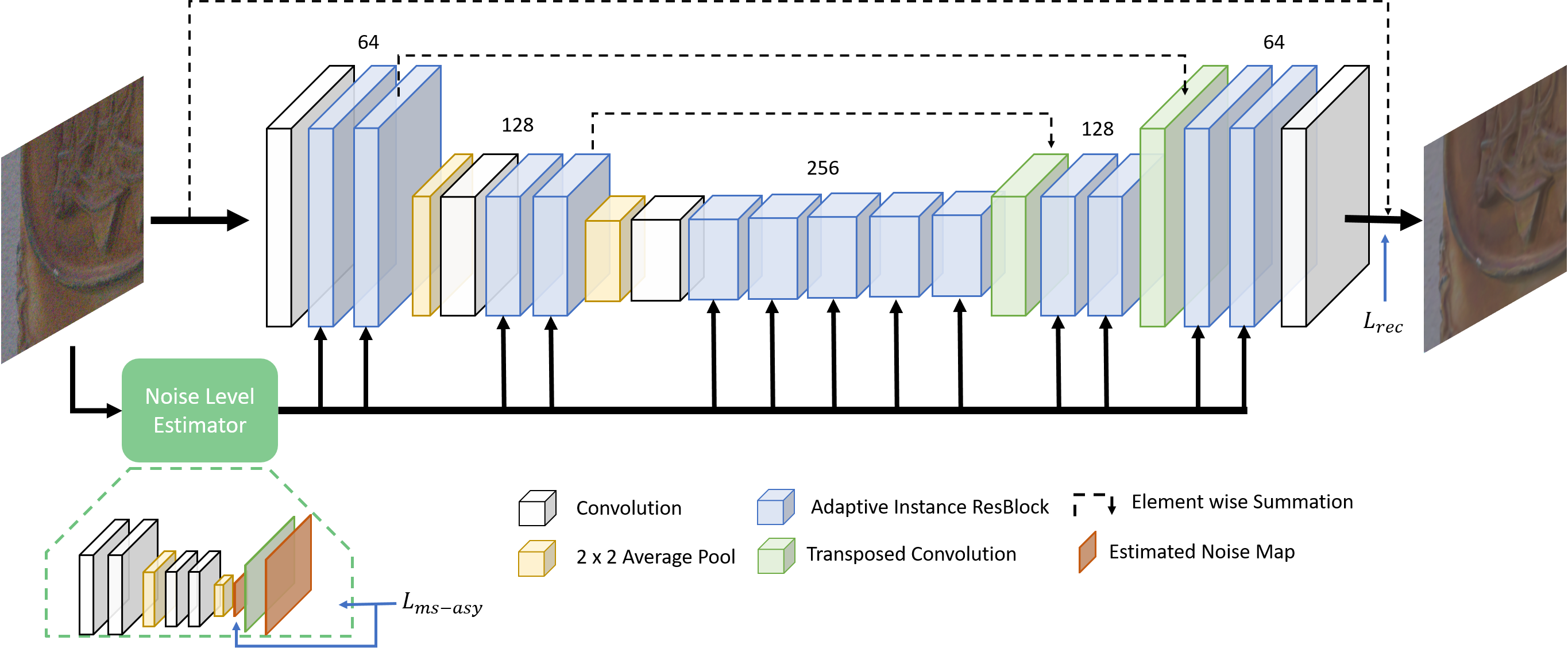} 
	\caption{
		Illustration of the proposed denoiser. 
		The noise level estimator and reconstruction network are U-Net based architecture, so the feature maps are down/up-sampled by average-pool/transposed convolution. We denote each scale of feature map as $1/s$ where $s$ can be 1, 2, and 4. All the represented convolutions in reconstruction network are $3\times3$ kernel having 64$s$ feature maps excluding last convolution.  Feature representation of noise level estimator is also composed of $3\times3$ convolutions with 32 channels and noise level maps are achieved from $3\times3$ convolutions having 3 channel outputs. 
		The amount of overall parameters is 13.7 M.
	}			
	\label{fig:overview}	
\end{figure*}

\section{Related Works}
The statistics of RN in standard RGB (sRGB) images depend on the properties of camera sensors and in-camera pipelines.
Specifically, shot noise and readout noise are generated from the sensor, and the statistics of generated noise are changed according to the in-camera pipeline such as demosaicing, gamma correction, in-camera denoiser, white balancing, color correction, etc~\cite{ortiz2004radiometric}.
There have been several works to approximate the RN model, including Gaussian-Poisson~\cite{foi2008practical,liu2014practical}, heteroscedastic Gaussian~\cite{hasinoff2010noise}, Gaussian Mixture Model (GMM)~\cite{zhu2016noise}, and deep leaning based methods~\cite{chen2018image,abdelhamed2019noise}.
Considering the camera pipeline, CBDNet~\cite{guo2019toward} and Unprocessing~\cite{brooks2019unprocessing} also considered realistic noise models. Specifically, they obtained near-RN images by adding the heteroscedastic Gaussian noise to the pseudo-raw images and feeding them to the camera pipeline.
These methods can simulate more than 200 camera response functions, and thus generate noisy images having different characteristics.
Moreover, CBDNet is alternately trained with the RN and SN to overcome overfitting to the noise model.
We think the alternate training scheme would incur training instability due to different data distributions, and also cannot train quite different RN  effectively.
Thus, we introduce a new transfer learning scheme that can simply but effectively adapt SN denoiser to other RN ones by re-transforming the normalized feature map.

%\subsection{Denoising Benchmark with Real Image}
%There have been several researches~\cite{anaya2018renoir,plotz2017benchmarking,abdelhamed2018high} that try to acquire the pair of noisy images and near noise-free image.
%Among them, Darmstadt Noise Dataset (DND)~\cite{plotz2017benchmarking} and Smartphone Image Denoising Dataset (SIDD)~\cite{abdelhamed2018high} achieved the high quality near noise-free images by spatially aligning low/high-ISO images, and overcoming intensity changes.
%Moreover, the SIDD attempts to address lens motion, radial distortion, and saturated intensity.
%We think these datasets are very well-prepared for evaluating the performance of denoising algorithms on real noise, and also for training the denoiser. However, although they tried to include many kinds of cameras for generating the datasets, there are infinitely many different cameras with the noise characteristics different from the existing datasets.  
%As a result, the CNNs trained with the existing real noise datasets have the same problem as the ones trained with the synthetic noise, in that they will not work well for the noise from the cameras that are not in the list of the devices used for building the datasets.
%-------------------------------------------------------------------------

\section{Proposed}
We aim to train a robust RN denoiser, which reduces the discrepancy between the distributions of training and test sets, by proposing a novel denoiser and transfer learning.
Precisely, we propose denoising architecture using the AIN, which can be well generalized to RN images.
Also, we introduce a transfer learning scheme to reduce the remaining data discrepancy, which consists of two stages:
(1) training a denoiser with SN dataset $\mathcal{S} = \{X_s,Y_s\}$ and (2) transfer learning with RN dataset $\mathcal{T} = \{X_r,Y_r\}$, where $X$ and $Y$ are noise-free images and noisy images respectively, and the subscript $s$ is for SN and $r$ for RN. We use the noise model from CBDNet for generating $Y_s$ from $X_s$ with the noise level of  ${\sigma}(\textbf{y}_s)$ where $\textbf{y}_s \in Y_s$ denotes SN image.  
After training SN denoiser with $\mathcal{S}$, RN denoiser is trained with $\mathcal{T}$ (pairs of RN image $\textbf{y}_r \in Y_r$ and near noise-free image $\textbf{x}_r \in X_r$). 
In the transfer learning stage, domain-specific parameters are only updated to effectively preserve learned knowledge from SN data.

% After training an SN denoiser with $\mathcal{S}$, the domain-specific parameter of RN denoiser, which is initially transfered from SN denoiser, is updated
%with RN image $\textbf{y}_r \in Y_r$ and near noise-free image $\textbf{x}_r \in X_r$.

\subsection{Adaptive Instance Normalization Denoising Network}
We present a novel AIN denoising network (AINDNet), where the same architecture is employed both for SN and RN denoiser.
We compose AINDNet with a noise level estimator and a reconstruction network, which is presented in Fig.~\ref{fig:overview}.
The noise level estimator takes a noisy image $\textbf{y}$ as an input and generates the estimated noise level map
$\hat{\sigma}(\textbf{y}) = F_{est}(\textbf{y};\theta_{est})$ where $\theta_{est}$ denotes a training parameter of estimator.
The reconstruction network takes $\hat{\sigma}$($\textbf{y}$) and $\textbf{y}$ as input and generates denoised image $\hat{\textbf{x}} = F_{rec}(\textbf{y},\hat{\sigma}(\textbf{y});\theta_{rec})$ where $\theta_{rec}$ denotes a training parameter of reconstruction network.
The reconstruction network is U-Net based architecture with AIN Residual blocks (AIN-ResBlocks).

\begin{figure}[t]
	\centering				
	\includegraphics[width=0.7\columnwidth]{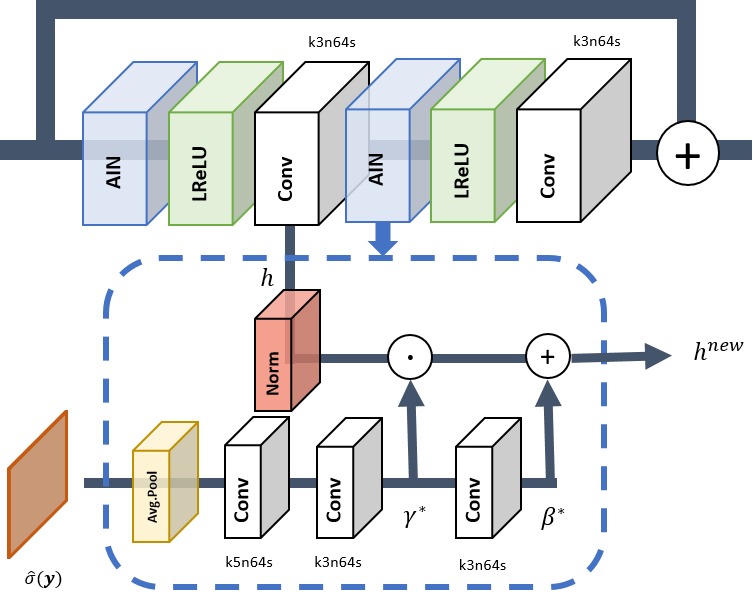} 
	\caption{Illustration of the proposed AIN-ResBlock with corresponding kernel size ($k$), feature scale ($s$), and number of features ($n$). Note that $n$ is linearly increasing according to $s$. Leaky ReLU is employed for an activation function. The Norm (red) block denotes channel-wise spatial normalization block. Average-pool scales the size of $\hat{\sigma}(\textbf{y})$ to be the same as that of $h$.}	
	\label{fig:AIN_Resblock}	
\end{figure}

\paragraph{Noise Level Estimator}
Estimating the noise level would not be an easy task due to the complex noise model and in-camera pipeline.
In our experiment, we find that previous simple noise level estimators~\cite{guo2019toward,brooks2019unprocessing}, which consist of five convolutions, could not accurately estimate the noise level.
The main reason is that the previous estimators have a small receptive field so that it could not fully capture complex noise information.
From this observation, we design a new noise level estimator with a larger receptive field by employing down/up-sampling and multi-scale estimations.
Specifically, estimator produces down-scaled estimation map $\hat{\sigma}_{4}(\textbf{y}) \in \!R^{H/4 \times W/4 \times 3} $ and original-sized estimation map $\hat{\sigma}_1(\textbf{y}) \in \!R^{H \times W \times 3} $.
Then, these two outputs are weight averaged to feed reconstruction network:
\begin{align}
\hat{\sigma}(\textbf{y}) &= \lambda_{ms} L(\hat{\sigma}_4(\textbf{y})) + (1-\lambda_{ms}) \hat{\sigma}_1(\textbf{y})
\label{eq:1}
\end{align}
where $H$, $W$, $L(\cdot)$ denotes the height and width of the image, and the linear interpolation respectively.
$\lambda_{ms}$ is empirically determined to 0.8.
From the weight average of multi-scale estimates, we can achieve region-wisely smoothed $\hat{\sigma}(\textbf{y})$, 
which follows general the characteristic of RN.

%We exclude total variation loss for training the noise level estimator, because $\mathcal{L}_{ms\text{-}asymm}$ can infer estimation map smoothly.  
\paragraph{Adaptive Instance Normalization}
The proposed AIN-ResBlock plays two crucial roles in the proposed denoising scheme.
One is regularizing the network not to be overfitted to SN images, and the other is adapting SN denoiser to RN denoiser.
For this, we build AIN-ResBlcok with two convolutions and two AIN modules, which is presented in Fig.~\ref{fig:AIN_Resblock}. 
The AIN module affine transforms normalized feature map $h\in \!R^{H' \times W' \times C}$ of convolution by taking a conditional input $\hat{\sigma}(\textbf{y})$ where $H' \times W'$ denotes the spatial size of feature map at each scale $s$, and $C$ is the number of channels.
Specifically, the AIN module produces affine transform parameters such as scale ($\gamma$) and shift ($\beta$) for each pixel.
Thus, every feature map is channel-wisely normalized and pixel-wisely affine transformed according to the noise level.
The update process of feature map in AIN module at site ($p \in \!R^{H'}$, $q \in \!R^{W'}$, $c \in \!R^{C}$) is formally represented as
\begin{align}
h^{new}_{p,q,c} &= \gamma^*_{p,q,c} \left(\frac{h_{p,q,c}-\mu_c}{\sigma_c} \right) + \beta^*_{p,q,c}
\label{eq:2}
\end{align}
where the variables with superscript * are generated from $\hat{\sigma}(\textbf{y})$, and $\mu_c$ and $\sigma_c$ denote the mean and standard deviation of $h$ respectively, in channel $c$. Precisely,
\begin{align}
\mu_c &= \frac{1}{H'W'} \sum_{p}^{H'} \sum_{q}^{W'} h_{p,q,c} \\
\sigma_c^2 &= \frac{1}{H'W'} \sum_{p}^{H'} \sum_{q}^{W'}(h_{p,q,c} - \mu_c)^2 + \epsilon
\end{align}  
where $\epsilon$ denotes the stability parameter, which prevents divide-by-zero in eq. (\ref{eq:2}), and we set $\epsilon=10^{-5}$ in our implementation. Note that
$\gamma^*_{p,q,c}$ and $\beta^*_{p,q,c}$ can be generated pixel-wisely and thus the proposed method can process spatially variant noisy images adaptively. 
In another point of view, AIN module acts as feature attention~\cite{chen2017sca,zhang2018image,woo2018cbam, dai2019second,kim2020agarnet} with explicitly constrained information ($\hat{\sigma}(\textbf{y})$).

\subsection{Transfer Learning}
\begin{figure}[t]
	\centering				
	\includegraphics[width=0.8\columnwidth]{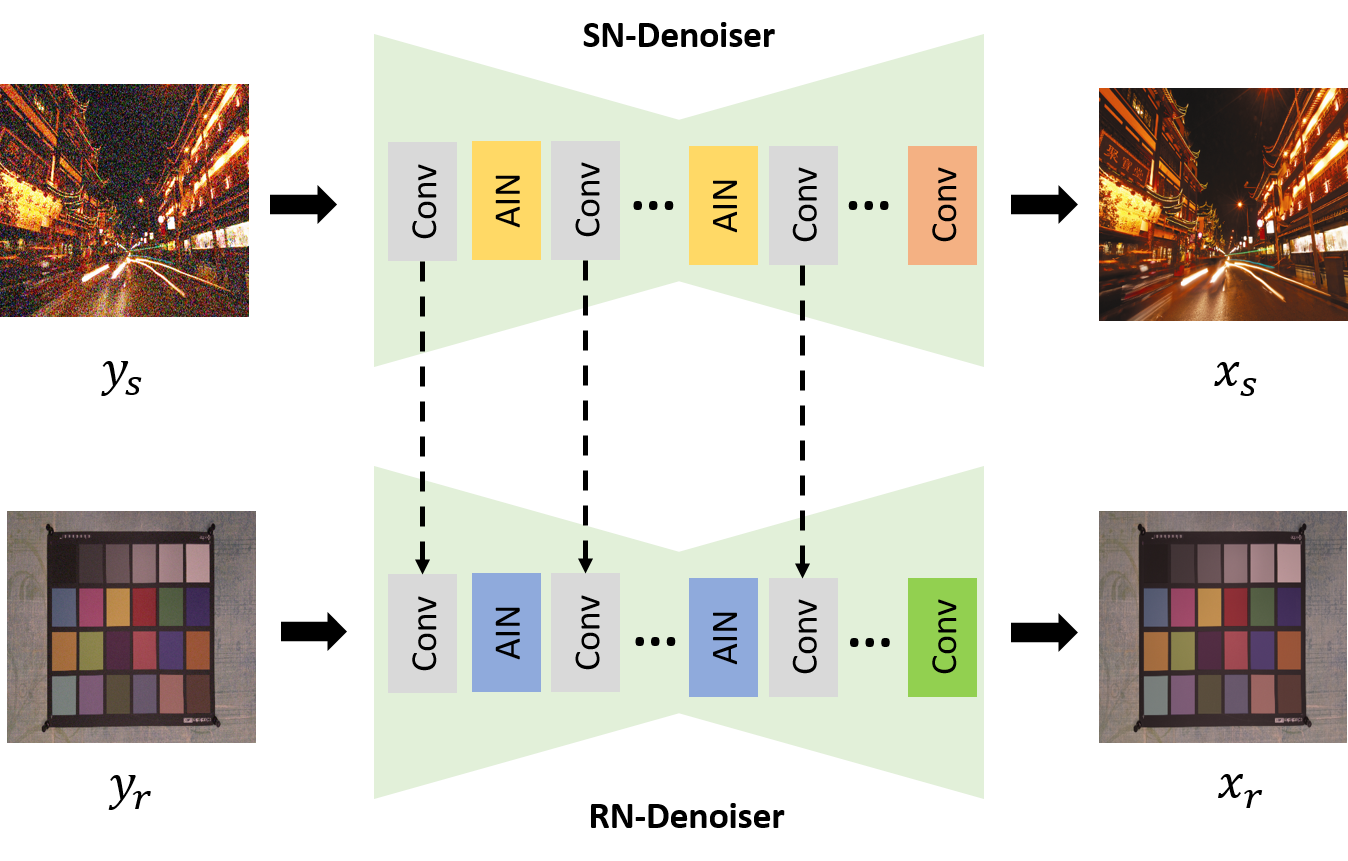} 
	\caption{Illustration of the proposed transfer learning scheme. 
		AIN module, noise level estimator, and last convolution are only updated when learning RN data.
		For the better visualization, we omit the noise level estimator in this figure.}	
	\label{fig:transfer_learning}	
\end{figure}

We propose transfer learning scheme to leverage $\mathcal{S}$ to accelerate the training of RN denoiser with $\mathcal{T}$ that has a limited number of elements (RN pairs). 
We expect that SN denoiser learns general and invariant feature representations  
and RN denoiser learns noise characteristics that cannot be fully modeled from SN data.
The proposed transfer learning scheme can achieve these two merits by adapting SN denoiser to RN denoiser.
For this, we focus on normalization parameter to handle different data distribution, which is inspired from other style transfer and classification tasks \cite{ulyanov2016instance,huang2017arbitrary, park2019semantic}. In these methods, transforming normalization parameters can transfer different style domain, and different domain classifications can be handled by switching the batch normalization parameters.
From these observations, we try to adapt different domain denoisers by transfer-learning the normalization parameters assuming that data discrepancy between  $\mathcal{S}$ and $\mathcal{T}$ can be adapted by re-transforming the normalized feature maps.

Specifically, AIN parameters of SN denoiser can be adapted pixel-wisely with conditional $\hat{\sigma}_i(\textbf{y}_s)$.
Thus, AIN modules and noise level estimator are transfer-learned with RN data.
Although the objective function of noise level cannot be present in $\mathcal{T}$, noise level estimator can be trained with the reconstruction loss.
We consider that last convolution plays a crucial role reconstructing feature maps to RGB image, hence last convolution is also updated. 
The overall proposed transfer learning scheme is presented in Fig.~\ref{fig:transfer_learning}.

Since the proposed transfer learning scheme only updates the parts of well generalized denoiser, it can be converged with faster speed and get better performance with very few number of elements from $\mathcal{T}$ than training from scratch. 
Moreover, the proposed scheme effectively copes with multiple models, which are inevitably required due to severely different noise statics, saving lots of memory by switching specific parameters.

%Lastly, proposed transfer learning scheme is easy to apply other low-level vision tasks when processing cross over domains.    

\paragraph{Training}
For training SN denoiser, we exploit multi-scale asymmetric loss as an estimation loss where asymmetric loss is introduced from CBDNet~\cite{guo2019toward} to prevent under estimation.
Formally, multi-scale asymmetric loss is defined as,
\begin{align}
\mathcal{L}_{ms\text{-}asymm} &= 
\sum_{i\in\{1,4\}}w_i |\alpha -  \mathbbm{1}_{(\hat{\sigma}_i(\textbf{y}_s) - \sigma_i(\textbf{y}_s) < 0)}| \\ &\cdot(\hat{\sigma}_i(\textbf{y}_s)-\sigma_i(\textbf{y}_s) ).^2 \nonumber
\end{align}
where $\mathbbm{1}$, $\cdot$, and $.^2$ denote element-wise operations such as indicator function, multiplication, and power respectively.
Hyperparameters $\{w_1, w_4, \alpha\}$ are empirically determined as $\{0.2, 0.8, 0.25\}$.
$\sigma_4(\textbf{y}_s)$ is achieved from $4 \times 4$ average pooling $\sigma_1(\textbf{y}_s)$.

Then, the proposed SN denoiser is jointly trained with estimation loss and $L_1$ reconstruction loss as,
\begin{align}
\mathcal{L} &= \|F(\textbf{y}_s;\theta_s) - \textbf{x}_s\|^1_1 + \lambda_{ms\text{-}asymm}\mathcal{L}_{ms\text{-}asymm}  
\end{align}
where $\theta_s$ denotes the SN denoiser training parameter including noise level estimator and reconstruction network.
$\lambda_{ms\text{-}asymm}$ denotes the weight term of noise level estimator and is empirically determined to 0.05.

For the RN denoiser, it is only trained with reconstruction loss:
\begin{align}
	\mathcal{L} &= \| F(\textbf{y}_r;\theta_r) - \textbf{x}_r \|^1_1 
\end{align}
where $\theta_r$ denotes the RN denoiser training parameter that is transferred from $\theta_s$. 
Previously stated parameter such as AIN modules, estimator, and last convolution are only updated, and other parameters are fixed when training the RN denoiser.
We use Adam optimizer for both SN denoiser and RN denoiser.

\section{Experiments}
We present the results of AWGN and RN images by training a Gaussian denoiser and RN denoiser.
 
\subsection{Experimental Setup}
\paragraph{Training Settings}
For the Gaussian denoiser, the training images are obtained from DIV2K~\cite{timofte2017ntire} and BSD400~\cite{martin2001database}, and noisy image is generated by AWGN model.
For the RN denoiser, we train a denoiser with two step: training an SN denoiser and training the RN denoiser by transfer learning. 
We achieve pairs of SN images and noise-free images from Waterloo dataset~\cite{ma2016waterloo} with heteroscedastic Gaussian noise model and simulating in-camera pipelines.
The RN denoiser, which is transferred from SN denoiser, is trained with SIDD training set~\cite{abdelhamed2018high}.
All the training images are cropped into patches of size $256 \times 256$.

\paragraph{Test Set}
In the AWGN experiments, we evaluate Set12~\cite{zhang2017beyond} and BSD68~\cite{roth2009fields} that are widely used for validating the AWGN denoiser.
Furthermore, we adopt three datasets for real-world noisy images:
\begin{itemize}
	\item RNI15~\cite{lebrun2015noise} is composed of 15 real-world noisy images. Unfortunately, the ground-truth clean images are unavailable, therefore we only present qualitative results.
	
	\item DND~\cite{plotz2017benchmarking} provides 50 noisy images that are captured by mirrorless cameras. 
	Since we cannot access near noise-free counterparts, the objective results (PSNR/SSIM) can be achieved by submitting the denoised images to DND site. 	
	
	\item SIDD~\cite{abdelhamed2018high} is obtained from smartphone cameras. 
	It provides 320 pairs of noisy images and corresponding near noise-free ones for the learning based methods where the captured scenes are mostly static.
	Furthermore, it provides 1280 patches for validation that has similar scenes with training set.
	The quantitative results (PSNR/SSIM) can be achieved by uploading the denoised image to SIDD site. 
\end{itemize}
 
\subsection{Comparison with state-of-the-arts}
\begin{table}[]
	\centering
	\caption{Average MAE and error STD for the images from Kodak24 where the inputs are corrupted by heteroscedastic Gaussian including in-camera pipeline. }
	\label{table:estimator}
	\begin{adjustbox}{width=0.6\linewidth}
		\begin{tabular}{l|cc|cc}
			\toprule
			Method   & \multicolumn{2}{c|}{FCN} & \multicolumn{2}{c}{Ours} \\
			\hline 
			($\sigma_s$, $\sigma_c$) & MAE        & STD        & MAE           & STD          \\
			\hline \hline
			(0.08, 0.02)     & 0.039      & 0.013      & \textbf{0.014}         & \textbf{0.012}        \\
			(0.08, 0.04)     & 0.059      & 0.014      & \textbf{0.012}         & \textbf{0.011}        \\
			(0.08, 0.06)     & 0.076      & 0.013      & \textbf{0.020}         & \textbf{0.010}        \\
			(0.12, 0.02)     & 0.052      & 0.021      & \textbf{0.015}         & \textbf{0.014}        \\
			(0.12, 0.04)     & 0.071      & 0.020      & \textbf{0.017}         & \textbf{0.014}        \\
			(0.12, 0.06)     & 0.087      & 0.020      & \textbf{0.030}         & \textbf{0.014}        \\
			\hline
			Average & 0.064 & 0.017& \textbf{0.018}& \textbf{0.013}\\
			\hline
			\# params & \multicolumn{2}{c|}{29.5 K} & \multicolumn{2}{c}{29.7 K} \\			
			\bottomrule			
		\end{tabular}
	\end{adjustbox}
\end{table}
\paragraph{Noise Level Estimation} 
We evaluate an accuracy of noise level estimator on exploited noise model images.
We compare the proposed noise level estimator with fully convolutional network (FCN) that are widely used~\cite{guo2019toward,brooks2019unprocessing}.
In order to evaluate the accuracy of estimator itself, each estimator is trained with $L_1$ regression.
The employed quantitative measurements are mean absolute error (MAE) and standard deviation (STD) of the error.
We report the accuracy of each estimator in Table~\ref{table:estimator} where the input images are simultaneously corrupted with signal dependent noise level $\sigma_s$ and signal independent noise level $\sigma_c$.
We can find that proposed estimator gets more accurate results than previous estimator with a similar number of parameters.
The results of more various noise levels will be presented in \textit{supplementary file}.
Furthermore, we will present the denoising performance when combined with reconstruction network.  
\begin{table}[]
	\centering
	\caption{Average PSNR of the denoised images, where the inputs are corrupted by AWGN with $\sigma = 15, 25,$ and $50$, for the images from Set12 and BSD68 datasets. (\color{red}{red}: \color{black} the best result, \color{blue}{blue}: \color{black} the second best)}
	\label{table:AWGN}
	\begin{adjustbox}{width = 0.85\linewidth }
		\begin{tabular}{l|ccc|ccc}
			\toprule
			Test Set     & \multicolumn{3}{c|}{Set12} & \multicolumn{3}{c}{BSD68} \\
			\hline 
			Method   & 15      & 25     & 50     & 15      & 25     & 50     \\
			\hline \hline
			BM3D~\cite{dabov2007color}    & 32.38   & 29.95  & 26.70  & 31.07   & 28.56  & 25.62  \\
			TNRD~\cite{chen2016trainable}    & 32.50   & 30.04  & 26.78  & 31.42   & 28.91  & 25.96  \\
			DnCNN~\cite{zhang2017beyond}  & 32.68   & 30.36  & 27.21  & 31.61   & 29.16  & 26.23  \\
			UNLNet~\cite{lefkimmiatis2018universal}  & 32.67   & 30.25  & 27.04  & 31.47   & 28.98  & 26.04  \\
			FFDNET~\cite{zhang2018ffdnet}  & 32.75   & 30.43  & 27.32  & 31.63   & 29.19  & 26.29  \\
			RIDNet~\cite{Anwar_2019_ICCV}  & \color{blue}{32.91}   & \color{blue}{30.60}  & \color{blue}{27.43}  & \color{red}{31.81}   & \color{red}{29.34}  & \color{red}{26.40}  \\
			AINDNet & \color{red}{32.92}   & \color{red}{30.61}  & \color{red}{27.51}  & \color{blue}{31.69}   & \color{blue}{29.26}  & \color{blue}{26.32}  \\		
			\bottomrule
		\end{tabular}
	\end{adjustbox}
\end{table}
\paragraph{AWGN Denoising}
We compare proposed denoiser on the noisy grayscale images that are corrupted by AWGN.
For this, we train Gaussian denoiser in a single network that learns noise level in [0,60].
The comparisons between the proposed method and other methods are presented in Table~\ref{table:AWGN}.
We can see that the proposed denoiser achieves the best performance on Set12 where composition of Set12 is independent from training sets.
On the other hand, the proposed method gets second best performance on BSD68 that consists of similar objects in BSD400 (training set).  
We think these results present robust generalization ability of the proposed denoising architecture for training set.

\paragraph{Real Noise Denoising}
We also investigate the proposed denoiser and transfer learning scheme on RN datasets.
Processing RN image is considered very practical, but difficult, because the noises are signal dependent, spatially variant, and visualized diversely according to different in-camera pipelines. Thus, we think RN denoising is an appropriate task for showing the generalization ability of the proposed denoiser and the effects of the proposed transfer learning. 

For the precise comparison, we train four different denoisers according to training sets and learning methods:
\begin{itemize}
	\item AINDNet(S): AINDNet is trained with SN images, which is proposed SN denoiser.
	\item AINDNet(R): AINDNet is trained with RN images.	
	\item AINDNet+RT: All the parameters from AINDNet(S) are re-trained with RN images, which is common transfer learning scheme.		
	\item AINDNet+TF: Specified parameters from AINDNet(S) are updated with RN images, which is proposed RN denoiser.	
\end{itemize}
Moreover, we present the geometric self-ensemble~\cite{timofte2016seven} results denoting super script $^*$ in order to maximize potential performance of the proposed methods.

Meanwhile, there have been a challenge on real image denoising~\cite{abdelhamed2019ntire} where the SIDD is used.
Our method shows lower performance than the top-ranked ones in the challenge, but it needs to be noted that the number of parameters of our network is much smaller than those in the challenge.
For example, DHDN~\cite{park2019densely} and DIDN~\cite{yu2019deep} that appeared in the challenge require about 160 M and 190 M training parameters respectively which are about 12 $\text{-}$ 15 times larger than ours.
Moreover, challenge methods have been slightly overfitted to SIDD where the winning denoiser~\cite{kim2019grdn} gets comparably lower performance (38.78 dB) on DND than our method. Therefore, we would not directly compare the proposed method with challenge methods.
 
The comparisons, including internal comparisons, are presented in Table~\ref{talbe:DND} and~\ref{talbe:SIDD}.
We can find that proposed methods get the best performance on DND and SIDD benchmarks.
Specifically, the proposed AINDNet(S) achieves the best performance on DND benchmark, 
which is impressive performance that outperforms RN trained denoisers. 
Moreover, AINDNet(S) gets 1.5 dB and 2.4 dB gains from CBDNet on DND and SIDD respectively where employed noise models are the same.
These results indicate that the proposed denoiser is not overfitted to noise model and can be well generalized to RN images.
However, AINDNet(S) has inferior performance than AINDNet(R) on SIDD with big margin. 
The main reason is that AINDNet(R) is solely trained with SIDD training images where test set consists of similar scenes and objects in training set.
In other words, AINDNet(R) can be slightly overfitted to SIDD benchmark and this phenomenon can be seen from insufficient performance on DND.  

In contrast, AINDNet+RT and AINDNet+TF get satisfying performance on both DND and SIDD.
Concretely, AINDNet+RT and AINDNet+TF have better performance than others, including AINDNet(R) on SIDD, 
which indicates that pre-training the SN images results in better performance.
AINDNet+TF more likely preserves priorly learned knowledges from SN data than AINDNet+RT, so AINDNet+TF achieves the best overall performance among compared methods.

\begin{figure}[t]
	\centering	
	\begin{subfigure}[b]{0.30\linewidth}		
		\includegraphics[width=1\columnwidth]{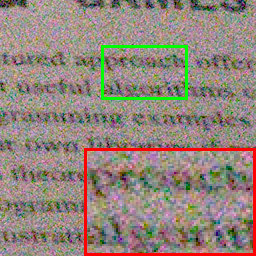} 
		\caption{Noisy Image}		
	\end{subfigure}	
	\begin{subfigure}[b]{0.30\linewidth}		
		\includegraphics[width=1\columnwidth]{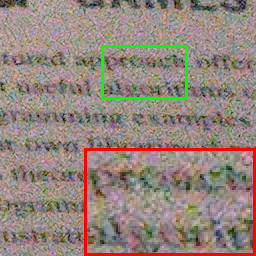}
		\caption{DnCNN-C}		
	\end{subfigure}	
	\begin{subfigure}[b]{0.30\linewidth}		
		\includegraphics[width=1\columnwidth]{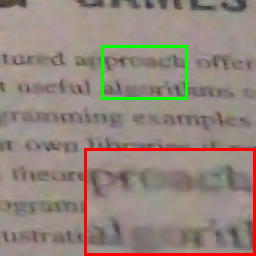}
		\caption{CBDNet}		
	\end{subfigure}	
	\begin{subfigure}[b]{0.30\linewidth}		
		\includegraphics[width=1\columnwidth]{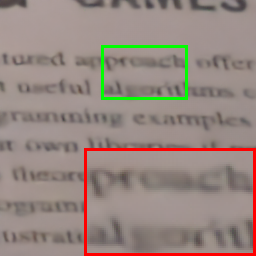} 
		\caption{RIDNet}		
	\end{subfigure}	
	\begin{subfigure}[b]{0.30\linewidth}		
		\includegraphics[width=1\columnwidth]{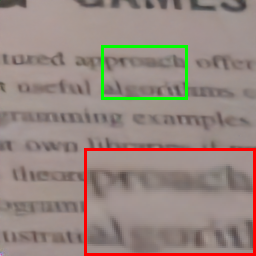}
		\caption{AINDNet(S)}		
	\end{subfigure}	
	\begin{subfigure}[b]{0.30\linewidth}		
		\includegraphics[width=1\columnwidth]{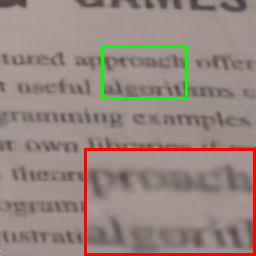}
		\caption{AINDNet+TF}		
	\end{subfigure}	
	\caption{The real noisy image from SIDD, and the comparison of the results.}
	\label{fig:comparison3}

	\begin{subfigure}[b]{0.30\linewidth}		
		\includegraphics[width=1\columnwidth]{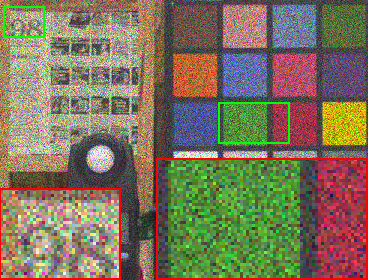} 
		\caption{Noisy Image}		
	\end{subfigure}	
	\begin{subfigure}[b]{0.30\linewidth}		
		\includegraphics[width=1\columnwidth]{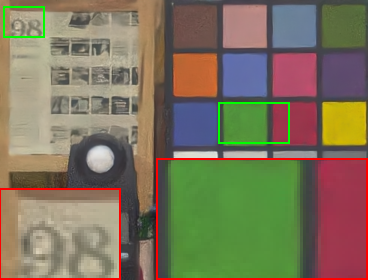}
		\caption{DnCNN-C}		
	\end{subfigure}	
	\begin{subfigure}[b]{0.30\linewidth}		
		\includegraphics[width=1\columnwidth]{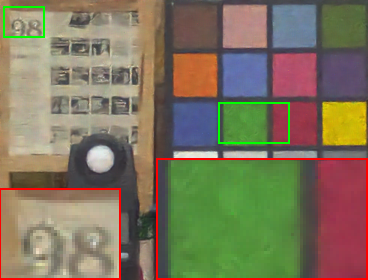}
		\caption{CBDNet}		
	\end{subfigure}	
	\begin{subfigure}[b]{0.30\linewidth}		
		\includegraphics[width=1\columnwidth]{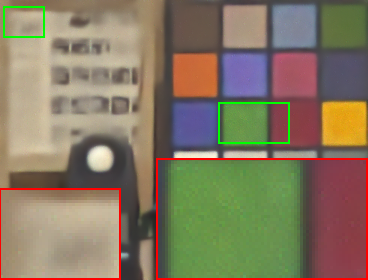} 
		\caption{RIDNet}		
	\end{subfigure}	
	\begin{subfigure}[b]{0.30\linewidth}		
		\includegraphics[width=1\columnwidth]{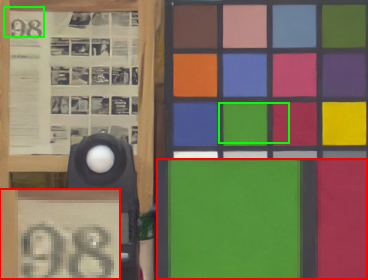}
		\caption{AINDNet(S)}		
	\end{subfigure}	
	\begin{subfigure}[b]{0.30\linewidth}		
		\includegraphics[width=1\columnwidth]{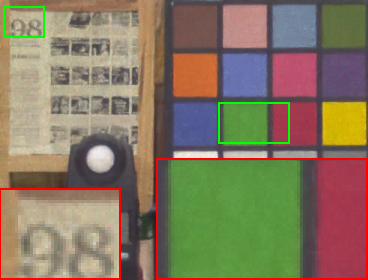}
		\caption{AINDNet+TF}		
	\end{subfigure}	
	\caption{The real noisy image from RNI15, and the comparison of the results.}
	\label{fig:comparison}	
\end{figure}

\begin{figure}[t]
	\centering	
	\begin{subfigure}[b]{0.31\linewidth}		
		\includegraphics[width=1\columnwidth]{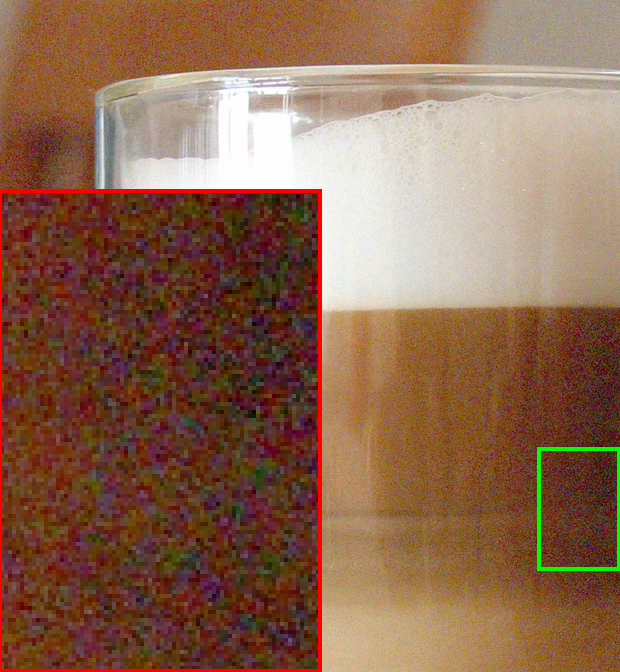}
		\caption{Noisy Image}		
	\end{subfigure}	
	\begin{subfigure}[b]{0.31\linewidth}		
		\includegraphics[width=1\columnwidth]{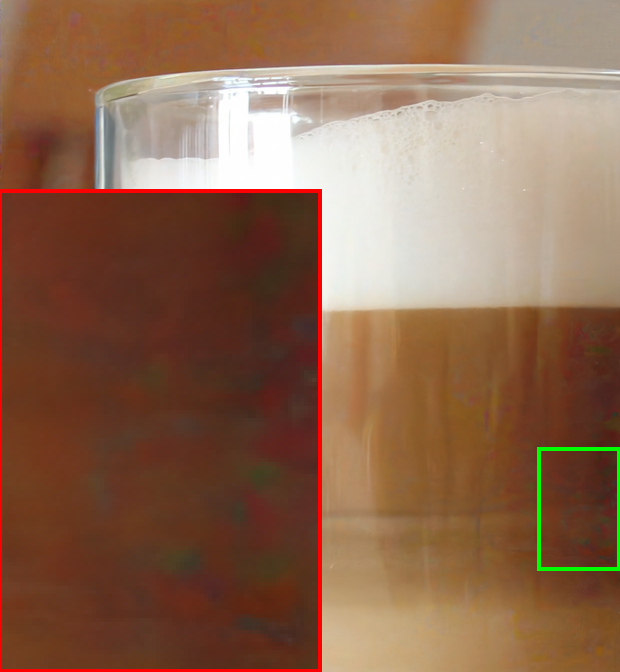}
		\caption{AINDNet(S)}		
	\end{subfigure}	
	\begin{subfigure}[b]{0.31\linewidth}		
		\includegraphics[width=1\columnwidth]{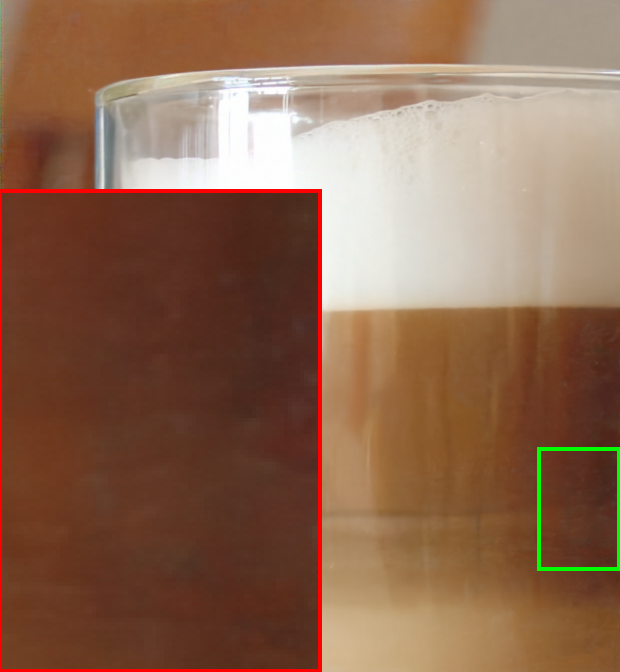} 
		\caption{AINDNet+TF}		
	\end{subfigure}		
	\begin{subfigure}[b]{0.31\linewidth}		
		\includegraphics[width=1\columnwidth]{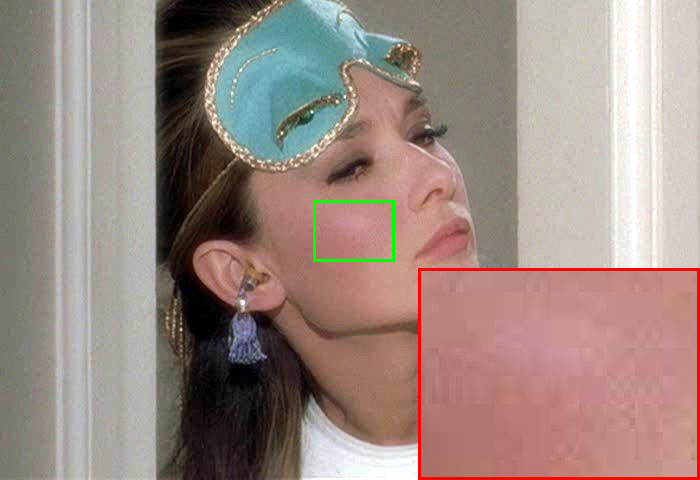} 
		\caption{Noisy Image}		
	\end{subfigure}	
	\begin{subfigure}[b]{0.31\linewidth}		
		\includegraphics[width=1\columnwidth]{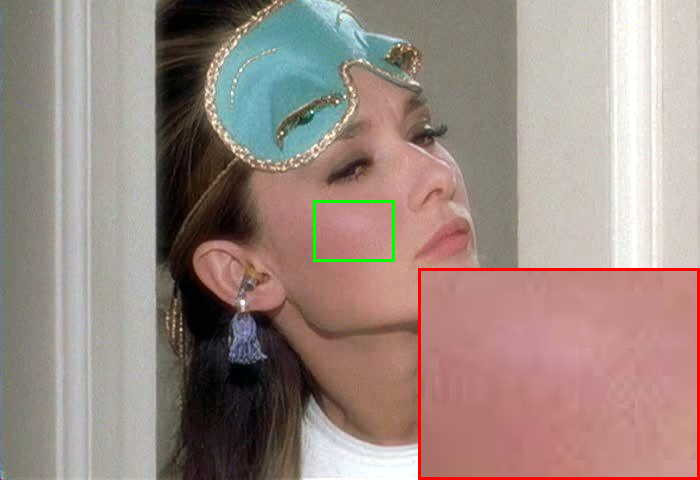} 
		\caption{AINDNet(S)}		
	\end{subfigure}	
	\begin{subfigure}[b]{0.31\linewidth}		
		\includegraphics[width=1\columnwidth]{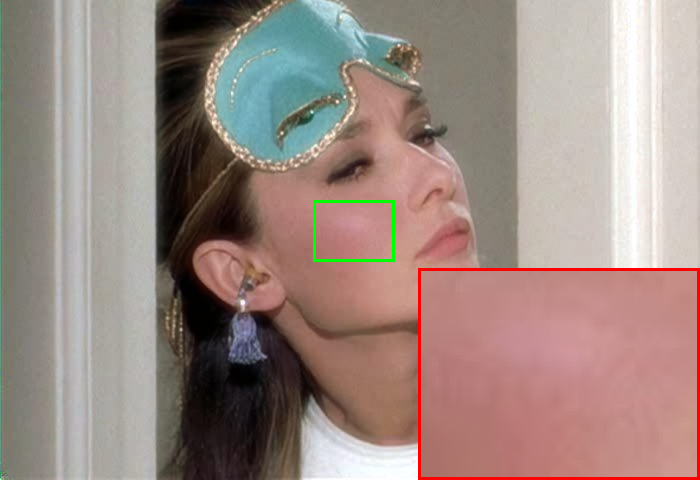}
		\caption{AINDNet+TF}		
	\end{subfigure}		
	\caption{The real noisy image from RNI15, and the comparison of the results showing the effectiveness of the proposed transfer learning scheme.}
	\label{fig:comparison2}
\end{figure}

We present visualized comparisons on SIDD and RNI15 in Figs.~\ref{fig:comparison3} and \ref{fig:comparison}, which show that proposed methods remove noises robustly while preserving the edges. 
Thus, characters in output images are more apparent than in other methods' results.
Furthermore, we also present visual enhancement in Fig.~\ref{fig:comparison2} when the proposed transfer learning scheme is applied.
Since RN denoiser transfer-learns characteristics of RN, AINDNet+TF successfully removes unusual noise that cannot be removed with AINDNet(S).
Moreover, RN denoiser learns the properties of JPEG compression artifacts that is not priorly learned in SN denoiser, so it can also successfully reduces compression artifacts. 
We will also present other visualized comparisons in \textit{supplementary file}.

\begin{table}[t]
	\centering
	\caption{Average PSNR of the denoised images on the DND benchmark, we denote the environment of training, {\em i.e.}, training with SN data only, RN data only, and both. $^*$ denotes geometric self-ensemble~\cite{timofte2016seven} result. (\color{red}{red}: \color{black} the best result, \color{blue}{blue}: \color{black} the second best)}
	\label{talbe:DND}
	\begin{adjustbox}{width=0.94\linewidth}
		\begin{tabular}{lllll}
			\toprule			
			Method & Blind/Non-blind   & Training Env.    & PSNR & SSIM  \\
			\midrule
			CDnCNN-B~\cite{zhang2017beyond}         & Blind 	 & Synthetic  & 32.43      & 0.7900      \\
			TNRD~\cite{chen2016trainable}         	 & Non-blind & Synthetic  & 33.65      & 0.8306      \\
			MLP~\cite{burger2012image}         	 & Non-blind & Synthetic  & 34.23  	  & 0.8331      \\
			FFDNet~\cite{zhang2018ffdnet}         	 & Non-blind & Synthetic  & 34.40  	  & 0.8474      \\
			BM3D~\cite{dabov2007color}         	 & Non-blind & -  & 34.51  	  & 0.8507      \\
			WNNM~\cite{gu2014weighted}         	 & Non-blind & -  & 34.67  	  & 0.8646      \\
			GCBD~\cite{chen2018image}         	 & Blind 	 & Synthetic  & 35.58  	  & 0.9217      \\
			KSVD~\cite{aharon2006k}         	 & Non-blind & -  & 36.49  	  & 0.8978      \\
			TWSC~\cite{xu2018trilateral}         	 & Blind 	 & -  & 37.94 	  & 0.9403      \\
			CBDNet~\cite{guo2019toward}    	 & Blind 	 & Synthetic  & 37.57  	  & 0.9360      \\
			CBDNet~\cite{guo2019toward}     & Blind 	 & Real  & 37.72  	  & 0.9408      \\
			CBDNet~\cite{guo2019toward}    	 & Blind 	 & All  & 38.06  	  & 0.9421      \\				
			RIDNet~\cite{Anwar_2019_ICCV}					& Blind		& Real	& 39.23 & 0.9526 \\		
			\midrule
			AINDNet(S) 	    	& Blind 	 & Synthetic  & \color{blue}{39.53}  	  & \color{blue}{0.9561}      \\	
			AINDNet(R) 		& Blind 	 & Real  & 39.16  	  & 0.9515		\\
			AIDNet + RT			& Blind 	 & All  & 39.21  	  & 0.9505		\\
			AINDNet + TF 		& Blind 	 & All  & 39.37  	  & 0.9505 		\\
			\midrule
			AINDNet(S)$^*$ 	    	 & Blind 	 & Synthetic  & \color{red}{39.77}  	  & \color{red}{0.9590} \\
			AINDNet(R)$^*$     	 & Blind 	 & Real  & 39.34  	  & 0.9524 \\
			AINDNet + RT$^*$ 			 & Blind 	 & All  & 39.34  	  & 0.9522 \\
			AINDNet + TF$^*$ 	    	 & Blind 	 & All  & 39.52  	  & 0.9522 \\
			\bottomrule					
		\end{tabular}
	\end{adjustbox}
\end{table}

\begin{table}[t]
	\centering
	\caption{Average PSNR of the denoised images on the SIDD benchmark, we denote the environment of training, {\em i.e.}, training with SN data only, RN data only, and both. $^*$ denotes geometric self-ensemble~\cite{timofte2016seven} result. (\color{red}{red}: \color{black} the best result, \color{blue}{blue}: \color{black} the second best)}
	\label{talbe:SIDD}
	\begin{adjustbox}{width=0.88\linewidth}
		\begin{tabular}{lllll}
			\toprule			
			Method & Blind/Non-blind   & Training Env.   & PSNR & SSIM  \\
			\midrule
			CDnCNN-B~\cite{zhang2017beyond}         & Blind 	 & Synthetic  & 23.66      & 0.583      \\		
			MLP~\cite{burger2012image}         	 & Non-blind & Synthetic  & 24.71  	  & 0.641      \\	
			TNRD~\cite{chen2016trainable}         	 & Non-blind & Synthetic  & 24.73      & 0.643      \\	
			BM3D~\cite{dabov2007color}         	 & Non-blind & -  & 25.65  	  & 0.685      \\			
			WNNM~\cite{gu2014weighted}         	 & Non-blind & -  & 25.78  	  & 0.809      \\			
			KSVD~\cite{aharon2006k}         	 & Non-blind & -  & 26.88  	  & 0.842      \\				
			CBDNet~\cite{guo2019toward}   	 & Blind 	 & All  & 33.28  	  & 0.868      \\			
			\midrule
			AINDNet(S) 	    	& Blind 	 & Synthetic  & 35.66  	  & 0.903      \\	
			AINDNet(R) 			& Blind 	 & Real  & 38.73  	  & 0.950		\\
			AIDNet + RT	 		& Blind 	 & All  & 39.04  	  & \color{red}{0.955}		\\
			AINDNet + TF		& Blind 	 & All  & 38.95  	  & 0.952 		\\
			\midrule
			AINDNet(S)$^*$ 	    	 & Blind 	 & Synthetic  & 35.87  	  & 0.905 \\
			AINDNet(R)$^*$	    	 & Blind 	 & Real  & 38.84  	  & 0.951 \\
			AINDNet + RT$^*$ 			 & Blind 	 & All  & \color{red}{39.15}  	  & \color{red}{0.955} \\
			AINDNet + TF$^*$ 	    	 & Blind 	 & All  & \color{blue}{39.08}  	  & \color{blue}{0.953} \\
			\bottomrule						
		\end{tabular}
	\end{adjustbox}
\end{table}

\begin{table*}[t]
	\centering
	\caption{Investigation of denoiser RN denoising performance according to the amount of RN dataset. The quantitative results (in average PSNR (dB)) are reported on SIDD validation dataset. }
	\label{table:Ablation2}
	\begin{adjustbox}{width=0.7\linewidth}
		\begin{tabular}{lccccccccc}
			\toprule			
			Num of Real Images &  0  & 1 & 2& 4 &8 &16 & 32 & 64 & 320 (full) \\
			\midrule
			RIDNet & - & - & - & - & - & - & - & - & 38.71  \\
			AINDNet(R) &  -  & 30.36 &	32.19 &	36.94 &	37.70 &	38.14 &	38.66 &	38.70 &	38.81 \\
			AINDNet+RT & 35.21 &	36.23 &	37.16 &	38.02 &	38.40 &	38.63 &	38.82 &	39.00 &	39.01  \\			
			AINDNet+TF & 35.21 &	36.19 &	37.14 &	37.93 &	38.27 &	38.52 &	38.75 &	38.83 &	38.90  \\
			
			\bottomrule					
		\end{tabular}
	\end{adjustbox}
\end{table*}

\subsection{Discussions}
\paragraph{Effect of Transfer Learning with Limited RN Pairs}
We investigate the relation between denoising performance and the amount of RN image pairs in $\mathcal{T}$, 
because we consider that preparation of $\mathcal{T}$ is quite difficult and the number of elements can also be limited. 
For this, we train each network with constrained image pairs from one to all (320) from SIDD~\cite{abdelhamed2018high}.
The average PSNR of each denoiser is presented in Table~\ref{table:Ablation2}. 
It can be seen that transfer learning schemes can infer great performance with the small number of real training images. 
It is notable that AINDNet+TF trained with 32 pairs of real data achieves better performance than RIDNet that exploits all.
Thus, we can conclude that the transfer learning with SN denoiser dramatically accelerate the performance with a small number of labeled data from other domain.

\paragraph{Architecture of Denoiser}
We demonstrate the effectiveness of reconstruction network for training with $\mathcal{S}$.
For this, AINDNet(S) is compared with a baseline (IN + Concat), which replaces AIN module with IN and concatenated input of noisy image and noise level map~\cite{zhang2017learning,zhang2018ffdnet}. 
Furthermore, we compare an adaptive Gaussian denoiser~\cite{kim2019adaptively} that can process spatially variant noise map by feeding gated-residual block (Gated-ResBlock).
Since it has not reported the performance on RN dataset, 
we train SN denoiser by replacing AIN-ResBlock to Gated-Resblock where other settings are same as AINDNet.
Table~\ref{table:reconstruction} shows that the proposed AIN-ResBlock shows the best performance on RN datasets.
Thus, we believe that the AIN-ResBlock is an appropriate architecture for the generalization.
We will present ablation study about update variable for transfer learning in \textit{supplementary} file.

\begin{table}[t]
	\centering	
	\caption{Investigation of the proposed reconstruction network when denoisers are trained with SN data. The quantitative results (in average PSNR (dB))  are reported on DND test dataset and SIDD validation dataset.}
	\label{table:reconstruction}
	\begin{adjustbox}{width=0.55\linewidth}
		\begin{tabular}{lcc}
			\toprule
			Method   & DND & SIDD \\
			\midrule			
			IN + Concat & 38.53 & 34.74 \\
			Res-Block~\cite{kim2019adaptively} & 39.19 & 34.93 \\
			Ours & \textbf{39.53} & \textbf{35.19} \\			
			\bottomrule			
		\end{tabular}
	\end{adjustbox}
\end{table}

\section{Conclusion}
In this paper, we have presented a novel denoiser and transfer learning scheme of RN denoising. The proposed denoiser employs an AIN to regularize the network and also to prevent the network from overfitting to SN. The transfer learning mainly updates the AIN module using RN data to adjust data distribution. From the experimental results, we could find that the proposed denoising scheme can be well generalized to RN even if it is trained with SN.
Moreover, the transfer learning scheme can effectively adapt an SN denoiser to an RN denoiser, with very few additional training with real- noise pairs. We will make our codes publicly available at https://github.com/terryoo/AINDNet for further research and comparison.

\paragraph{Acknowledgments}
This work was supported in part by Institute
for Information \& communications Technology Planning
\& Evaluation (IITP) grant funded by the Korea government(
MSIT) (No.NI190004,Development of AI based
Robot Technologies for Understanding Assembly Instruction
and Automatic Assembly Task Planning), and in part
by Samsung Electronics Co., Ltd.

{\small
\bibliographystyle{ieee_fullname}

}

\section{Transfer Learning from AWGN}
We present the results of transfer-learned denoiser where AINDNet is pre-trained with AWGN and adapted to real noise (RN).
For the precise comparison, we report performance of three denoisers in Table~\ref{talbe:AWGN} according to training sets and learning methods:
\begin{itemize}
	\item AINDNet(AWGN): AINDNet is trained with AWGN images.	
	\item AINDNet(AWGN)+TF$_1$: AINDNet(AWGN) is transfer learned with a single real noisy image.		
	\item AINDNet(AWGN)+TF: AINDNet(AWGN) is transfer learned with full real noisy images (320 images).	
\end{itemize}
It can be seen that proposed transfer learning scheme significantly improves the performance of synthetic noise (SN) denoisers including AWGN denoiser when the input is limited. 

\begin{table}[h]
	\centering
	\caption{Average PSNR of the denoised images on the SIDD validation set. $_1$ denotes that the number of real training noisy image is one.}
	\label{talbe:AWGN}
	\begin{adjustbox}{width=0.6\linewidth}
		\begin{tabular}{ll}
			\toprule			
			Method &   PSNR   \\
			\midrule
			RIDNet~\cite{Anwar_2019_ICCV} & 38.71 \\
			\midrule
			AINDNet(S) & 35.21  \\
			AINDNet(AWGN) & 26.25 \\
			AINDNet(R) & 38.81 \\
			AINDNet(AWGN)+TF &  38.82  \\
			AINDNet+TF & 38.90\\			
			\midrule
			AINDNet(R)$_1$ & 30.36 \\
			AINDNet(AWGN)+TF$_1$ & 31.76 \\
			AINDNet+TF$_1$ & 36.19 \\			
			\bottomrule						
		\end{tabular}
	\end{adjustbox}
\end{table}

\section{More Noise Level Estimation Results}
We evaluate the accuracy of the proposed noise level estimator, where the input images are simultaneously corrupted with more diverse signal-dependent noise levels $\sigma_s$ and signal-independent noise levels $\sigma_c$. As presented in Table~\ref{table:estimator_more}, the proposed noise level estimator achieves better accuracy with lower standard deviations of the errors in most cases.
Furthermore, the proposed noise level estimator predicts quite accurate estimates when the images are corrupted with high $\sigma_s$ and $\sigma_c$.

\begin{table}[h]
	\centering
	\caption{Average MAE and error STD for the images from Kodak24 where the inputs are corrupted by heteroscedastic Gaussian including in-camera pipeline. }
	\label{table:estimator_more}
	\begin{adjustbox}{width=0.8\linewidth}
		\begin{tabular}{l|cc|cc}
			\toprule
			Method   & \multicolumn{2}{c|}{FCN~\cite{guo2019toward}} & \multicolumn{2}{c}{Ours} \\
			\hline 
			($\sigma_s$, $\sigma_c$) & MAE        & STD        & MAE           & STD          \\
			\hline \hline
			(0.04,	0.00) &	\textbf{0.009} &	\textbf{0.007} &	0.022 &	0.014 \\
			(0.04,	0.02) &	0.029 &\textbf{0.007} &	\textbf{0.015} &	0.011 \\
			(0.04,	0.04) &	0.050 &	\textbf{0.006} &	\textbf{0.009} &	0.009 \\
			(0.04,	0.06) &	0.070 &	\textbf{0.007} &	\textbf{0.016} &	0.009 \\
			(0.08,	0.00) &	\textbf{0.018} &	\textbf{0.013} &	0.022 &	0.014 \\
			(0.08,	0.02) &	0.039 &	0.013 &	\textbf{0.014} &	\textbf{0.012} \\ 
			(0.08,	0.04) &	0.059 &	0.014 &	\textbf{0.012} &	\textbf{0.011} \\
			(0.08,	0.06) &	0.076 &	0.013 &	\textbf{0.020} &	\textbf{0.010} \\
			(0.12,	0.00) &	0.029 &	0.020 &	\textbf{0.020} &	\textbf{0.014} \\ 
			(0.12,	0.02) &	0.052 &	0.021 &	\textbf{0.015} &	\textbf{0.014} \\ 
			(0.12,	0.04) &	0.071 &	0.020 &	\textbf{0.017} &	\textbf{0.014} \\
			(0.12,	0.06) &	0.087 &	0.020 &	\textbf{0.030} &	\textbf{0.014} \\ 
			(0.16,	0.00) &	0.039 &	0.027 &	\textbf{0.021} &	\textbf{0.018} \\ 
			(0.16,	0.02) &	0.065 &	0.028 &	\textbf{0.020} &	\textbf{0.019} \\ 
			(0.16,	0.04) &	0.076 &	0.027 &	\textbf{0.021} &	\textbf{0.019} \\ 
			(0.16,	0.06) &	0.098 &	0.028 &	\textbf{0.040} &	\textbf{0.021} \\						
			\hline
			Average & 0.054 & 0.017& \textbf{0.020}& \textbf{0.014}\\
			\hline
			\# params & \multicolumn{2}{c|}{29.5 K} & \multicolumn{2}{c}{29.7 K} \\			
			\bottomrule			
		\end{tabular}
	\end{adjustbox}
\end{table}

\section{Ablation Study}
We demonstrate the effectiveness of noise level estimator for training with $\mathcal{S}$.
We present performance of noise level estimators combined with reconstruction network in Table~\ref{table:combined_estimator} with different objective function.
Remember that $L_{ms\text{-}asymm}$ can generate smoothed outputs, so $L_{TV}$ is excluded when using $L_{ms\text{-}asymm}$.
We find that state-of-the-art training scheme (FCN + $L_{asymm}$ + $L_{TV}$) infers inferior performance than proposed training scheme (Ours + $L_{ms\text{-}asymm}$).  
Moreover, the proposed training scheme also surpasses internal variation (Ours + $L_{1}$ + $L_{TV}$). 

\begin{table}[h]
	\centering
	\caption{Investigation of noise level estimator and estimation loss when denoisers are trained with SN data. The quantitative results (in average PSNR (dB)) are reported on DND test dataset and SIDD validation dataset.}
	\label{table:combined_estimator}
	\begin{adjustbox}{width=0.6\linewidth}
		\begin{tabular}{lcc}
			\toprule
			Method   & DND & SIDD \\
			\midrule
			FCN + $L_{asymm}$ + $L_{TV}$    & 39.51 & 34.90 \\		
			Ours + $L_{1}$ + $L_{TV}$    & 39.45 & 35.08 \\
			Ours + $L_{ms\text{-}asymm}$ & \textbf{39.53} & \textbf{35.19} \\ 		
			\bottomrule			
		\end{tabular}
	\end{adjustbox}
\end{table}

We further investigate the relation between update parameters and performance in the transfer learning phase.
For the precise comparison, we compare three variants by freezing each update parameter in Table~\ref{table:update_parameter}:
\begin{itemize}
	\item Ours\text{-}AIN: AIN module is not updated in transfer learning stage.	
	\item Ours\text{-}Estimator: Noise level estimator is not updated in transfer learning stage.
	\item Ours\text{-}LastConv: Last convolution is not updated in transfer learning stage.	
\end{itemize}
It can be seen that proposed updating the noise level estimator, and last convolution contribute 0.1 - 0.2 dB performance gain respectively. Fixing AIN module parameter presents even worse performance than the SN denoiser.

\begin{table}[h]
	\centering
	\caption{Investigation of update parameters when denoisers are transfer-learned with RN data. The quantitative results (in average PSNR (dB)) are reported on SIDD validation dataset.}
	\label{table:update_parameter}
	\begin{adjustbox}{width=0.4\linewidth}
		\begin{tabular}{lc}
			\toprule
			Method   &  PSNR \\
			\midrule			
			Ours\text{-}AIN    &  34.60 \\		
			Ours\text{-}Estimator   &  38.71 \\
			Ours\text{-}LastConv & 38.75 \\ 	
			\midrule
			AINDNet(S) & 35.21 \\
			AINDNet+TF & 38.90 \\	
			\bottomrule			
		\end{tabular}
	\end{adjustbox}
\end{table}

\end{document}